%% file: main.tex
\documentclass{article}
\usepackage{xcolor}
\usepackage[preprint]{corl_2026}
\usepackage{graphicx,booktabs,amsmath,microtype,longtable,array,placeins,etoolbox}
\usepackage[T1]{fontenc}
\graphicspath{{figures/}}
\definecolor{mover}{HTML}{147D92}
\definecolor{reference}{HTML}{C46635}
\title{Distilling Instructions into Weights:\\Closed-Form Edits Fix a Robot Policy's Misreading}
\author{Ali-Adeeb Abbas$^{1,2}$ and Esmaeil Seraj$^{2}$\thanks{$^1$RIVeR Laboratory, Northeastern University, Boston, MA, USA. $^2$General Motors, Mountain View, CA, USA. Correspondence: \texttt{abbas.a@northeastern.edu}. Project page: \url{https://ali-adeeb.com/weightsmith/}.}}
\hypersetup{pdftitle={Distilling Instructions into Weights: Closed-Form Edits Fix a Robot Policy's Misreading},pdfsubject={arXiv preprint},pdfauthor={Ali-Adeeb Abbas, Esmaeil Seraj}}

\makeatletter
\patchcmd{\@maketitle}{\vskip 0.1in}{\vskip 0pt}{}{\PackageError{preprint-layout}{Title top spacing patch failed}{}}
\patchcmd{\@maketitle}{\vskip 0.3in \@minus 0.1in}{\vskip 0.1in}{}{\PackageError{preprint-layout}{Title bottom spacing patch failed}{}}
\patchcmd{\@maketitle}{\bf\rule{\z@}{24\p@}\@author}{\bf\rule{\z@}{12\p@}\@author}{}{\PackageError{preprint-layout}{Author spacing patch failed}{}}
\patchcmd{\@maketitle}{\bf\rule{\z@}{24\p@}\@author}{\bf\rule{\z@}{12\p@}\@author}{}{\PackageError{preprint-layout}{Preprint author spacing patch failed}{}}
\makeatother

\newcommand{\pifive}{$\pi_{0.5}$}


\begin{document}
\maketitle
\begin{abstract}
Generalist robot policies often fail at one step of a task they can otherwise perform. We ask whether such a failure can be repaired by editing the policy's weights directly, without gradients, training data, or a change to the instruction. Told to put the cube then the banana in the bowl, a zero-shot \pifive{} vision-language-action policy in RoboLab releases the cube before it reaches the bowl. Replaying 21 of these decisions under rewordings shows that it binds ``in the bowl'' to the wrong object. Closed-form instruction distillation copies the per-layer effect of an explicit rewording into the language model's MLP weights. On 128 held-out seeds, success rises from 27 to 106 episodes, with 81 fixed and 2 broken ($p=7\times10^{-22}$), while an update of the same size in random directions leaves success unchanged. The edit also lifts two other tasks without detectable harm on the remaining 13, and a variant that stores 0.16\% of the parameter count keeps most of its effect. Placed in the action expert or action head instead, the same distillation helps less or not at all. Our released tool reproduced the fix end to end on new held-out seeds, from 8/64 to 36/64. Edits aimed at perception, a missing grasp, contact physics, or the gripper threshold did not help, and of seven blind Claude Opus editors, only the one given the misread instruction produced an edit that passed held-out scoring. These results suggest that weight edits work best when a known-good input already exists to distil. Because reruns with unchanged weights flip about a third of paired outcomes, evaluations should pair seeds, rerun the unedited policy, and confirm gains on held-out seeds.
\end{abstract}

\setlength{\parskip}{2.5pt}
\setlength{\intextsep}{8pt plus 2pt minus 2pt}
\setlength{\textfloatsep}{12pt plus 2pt minus 2pt}
\section{Introduction}
Vision-language-action (VLA) policies map camera images and an instruction to robot actions~\citep{pi2025pi05}. A generalist policy often fails at a single step of a task whose skills it already has, and the usual remedy is to collect demonstrations and fine-tune. Weight-space programming offers another route: a language model reads a network's weights and activations, locates the computation that goes wrong, and patches the weights directly, without gradients~\citep{takehi2026sparks}. Which failures of a robot policy can such a patch fix?

\begin{figure}[!t]
\centering
\includegraphics[width=\linewidth]{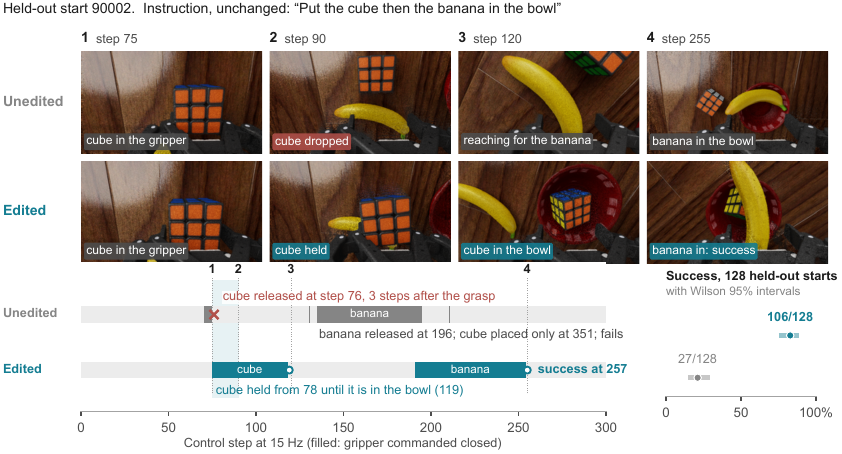}
\caption{One held-out start, unedited and edited. Top: wrist-camera inputs at four matched control steps. Bottom: commanded gripper (filled where closed) and logged events. Right: held-out success over 128 starts, with Wilson 95\% intervals. With the instruction unchanged, the edit removes the release three steps after the grasp. The explicit rewording ``Put the cube in the bowl then put the banana in the bowl'' was used only to build the edit.}
\label{fig:teaser}
\end{figure}

Early robot results are encouraging but hard to compare. Guo and Xie report a \pifive{} task rising from 28/92 to 82/92 held-out episodes after changing four parameters, on checkpoints fine-tuned for each task~\citep{guo2026robotweights,xie2026robotweights}. Gupta et al.\ report gains on five of ten \pifive{} tasks, and a folding gain from 51\% to 75\%, with an update built from the difference between two trained models~\citep{gupta2026robotweights}. In our companion study, four sparse edits that GPT-6 Astra wrote for a zero-shot \pifive{} stacking policy all failed a preservation gate~\citep{abbas2026wspstudy}. We did not reproduce these results. Together, they leave open which failures gradient-free edits can fix in a policy that was never trained on the task.

We study this question with the DROID version of \pifive{}~\citep{pi2025pi05,khazatsky2024droid} on 16 RoboLab tasks~\citep{yang2026robolab}, comparing every edit with the unedited policy on identical simulator seeds. Told to ``put the cube then the banana in the bowl,'' the policy grasps the cube and lets go before reaching the bowl, yet the same weights succeed under an explicit rewording. Robot policies are known to respond differently to equivalent wordings~\citep{fei2026liberoplus,kim2026liberopara,abbas2026samegoal}; here that sensitivity points to a repair. We introduce \emph{closed-form instruction distillation}, which finds such a failure by replaying recorded decisions under rewordings and then writes the internal effect of the working wording into the weights with a multi-key closed-form update~\citep{meng2023memit}. To our knowledge, it is the first edit that distils an instruction rewording into a robot policy in closed form. On a held-out start, the edited policy keeps hold of the cube that the unedited policy drops three control steps after grasping it (Fig.~\ref{fig:teaser}).

With the instruction unchanged, the edit raises held-out success from 27/128 to 106/128, while an update of the same size in random directions does nothing. A variant that stores 0.16\% of the parameter count keeps most of the effect, and the edit lifts two other tasks without detectable harm on the remaining 13. Where the edit is placed matters: written into the action expert or action head instead of the VLM, the same distillation helps less or not at all, and it no longer transfers. We release the procedure as a command, \texttt{weightsmith distill}; with decisions selected by rule rather than by hand, it raised held-out success on new seeds from 8/64 to 36/64. Edits aimed at perception, missing competence, contact physics, and gripper commitment did not help. Seven autonomous editors working blind, one per failure kind, confirm this pattern: only the one given the misread instruction produced an edit that passed held-out scoring, and a gated stacking edit removed the wrong decision it targeted without raising success. Finally, we measure the noise floor of paired closed-loop evaluation, where rerunning unchanged weights flips about a third of paired outcomes, and we release the protocol, code, and paired-episode data.

\section{Closed-Form Instruction Distillation}
\label{sec:method}
To repair a misreading without training, we first find an instruction that works and then write its effect into the weights. We replay recorded failure decisions under rewordings until one wording removes the failure, distil that wording into the weights in closed form, and check the edit offline before confirming it in closed loop, first on practice seeds and then on held-out seeds. Every step uses only weights, recorded inputs, and activations.

\pifive{} combines an 18-layer PaliGemma vision-language model (VLM) with an 18-layer Gemma action expert~\citep{pi2025pi05,beyer2024paligemma}. The VLM processes image tokens from two cameras together with the instruction and robot-state tokens. Its attention is bidirectional, so image tokens can attend to the instruction. The action expert attends to the VLM's keys and values and produces 15-step action chunks, including a gripper command, by flow matching~\citep{black2024pi0}.

\begin{figure}[!t]
\centering
\includegraphics[width=\linewidth]{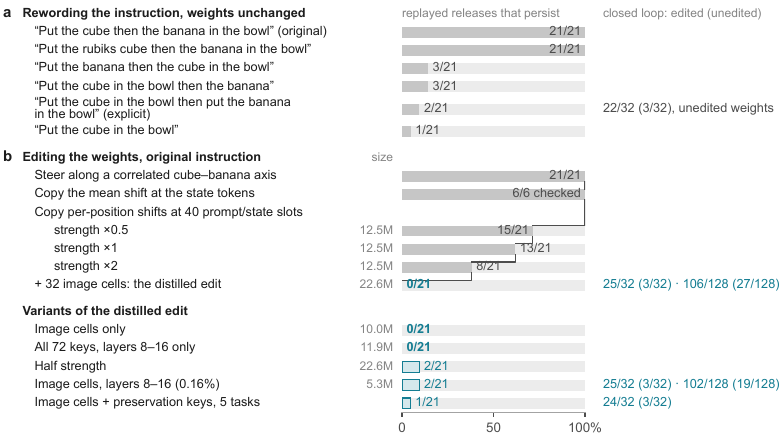}
\caption{What it takes to fix the decision. Bars give the share of 21 recorded premature-release decisions that still release when replayed offline with images, robot state, and noise fixed. (a) Rewordings. (b) Weight edits in the order they were built, then variants of the distilled edit. Size counts low-rank factors. Closed-loop results are edited (unedited) successes on practice $\cdot$ held-out seeds.}
\label{fig:construction}
\end{figure}

The cube-then-banana task illustrates the diagnosis. Under the original instruction the unedited policy succeeds in 9 of 64 episodes. It goes for the cube first in 63 of 64, grasps it, lifts it a few centimetres, and lets go well short of the bowl. We call this a \emph{premature release}. When the cube does reach the bowl first, the episode usually succeeds (about 7 times in 9), so the skill is present. We replayed 21 recorded premature-release decisions with images, robot state, and flow-matching noise fixed, changing only the instruction (Fig.~\ref{fig:construction}a). Renaming the cube leaves all 21 releases, whereas moving ``in the bowl'' next to the cube leaves only one to three. In one VLM attention head, the destination words also attend far more to the adjacent noun than to the first (Appendix~\ref{app:construction}). We read this as a binding error. While holding the cube, the policy treats the banana as the object that belongs in the bowl. We call the explicit wording a \emph{known-good instruction}. With unchanged weights it succeeds in 22 of 32 closed-loop episodes against 3 of 32 for the original on the same seeds (20 fixed, 1 broken).

Distillation makes the network process the original instruction as it processes the known-good one. At each VLM layer $\ell=0,\dots,16$ we form 72 keys, the rows of $K_\ell$. A key is the mean input to the layer's MLP down-projection under the original instruction, averaged over 6 recorded premature-release decisions. The keys sit at 40 positions shared by both prompts (the beginning-of-sequence, ``Task:'', and robot-state tokens) and at 32 image cells (16 per camera). The matching rows of $V_\ell$ hold the layer's residual increment at each position under the explicit instruction minus that under the original. With activations as row vectors, we apply the regularized minimum-norm update that maps each key to its value,
\begin{equation}
\Delta W_\ell = K_\ell^{\top}\left(K_\ell K_\ell^{\top} + \lambda I\right)^{-1} V_\ell,\qquad \lambda = 10^{-3}\,\mathrm{tr}\!\left(K_\ell K_\ell^{\top}\right)/72,
\label{eq:edit}
\end{equation}
so that $K_\ell \Delta W_\ell \approx V_\ell$. Eq.~\eqref{eq:edit} is the multi-key closed form used to rewrite stored associations in language models~\citep{meng2022rome,meng2023memit}. We compute it once per layer and aim it at a rewording. Context distillation trains a model to act as if a prompt were present~\citep{snell2022distilling}, whereas Eq.~\eqref{eq:edit} needs no training. The update has rank at most 72 per layer.

Offline, the edit removes all 21 premature releases. It leaves unchanged all 9 recorded correct carries and all 19 phase-two decisions, in which the policy fetches the banana after placing the cube. Its actions also land closer to the explicit wording's actions than the unedited ones do (4.5 versus 8.1\,cm at the end of the chunk). On the 32 practice seeds, closed-loop success rises from 3/32 to 25/32, with 22 episodes fixed and none broken (exact McNemar $p=4.8\times10^{-7}$). We then froze the edit for held-out evaluation.

\section{Paired Weight-Edit Evaluation}
\label{sec:eval}
We compare each edited policy with the unedited policy from identical simulator starts, using the OpenPI \pifive{} DROID joint-position checkpoint with about 3.35 billion parameters~\citep{openpi2025,pi2025pi05}. It was trained on DROID data~\citep{khazatsky2024droid} and on none of our tasks, so every result is zero-shot. It receives two $224\times224$ camera images and the joint and gripper state. It outputs 15-step chunks of seven joint commands and a gripper command, which is binarized at 0.5. We run 16 RoboLab tasks in Isaac Sim with 16 parallel environments per evaluation~\citep{yang2026robolab}. An episode counts as a success when RoboLab's terminal success flag is set.

A 16-episode survey of the unedited policy on each task sorted its failures into levels (Appendix~\ref{app:survey}, Table~\ref{tab:levels}). We call a failure \emph{language binding} when the wrong object counts as the current target, and \emph{perception} when the arm goes for a similar-looking object. \emph{Missing competence} means that no working motion is produced. In a \emph{contact physics} failure the decisions look right but the grasp slips, and in a \emph{gripper commitment} failure open and close decisions sit near the threshold. A motor decision that stalls inside a learned skill did not occur among our tasks.

Edits came from two frontier models run headless, Claude Opus in this work and GPT-6 Astra in the companion study~\citep{abbas2026wspstudy}. An editor can read and write every weight in a NumPy sandbox, capture activations on the exact inputs of any recorded decision, and roll the policy out. It may not use gradients, optimizers, fine-tuning, or black-box search. Every edit is a closed-form computation from weights and activations, and its hypothesis is written down before evaluation. The instruction text stays fixed. The operator session, also Claude Opus, built some edits under the same rules, including the distilled one. Seven autonomous editors repeat the protocol blind in Section~\ref{sec:blind}.

A simulator \emph{slot} fixes the random seeds and scene of one episode. Every comparison runs both policies on the same slots, and we call these \emph{paired seeds}. Each pair ends in one of four outcomes. Both policies succeed, both fail, only the edited policy succeeds, or only the unedited policy succeeds. We call the last two \emph{fixed} and \emph{broke}. Edits are developed on \emph{practice seeds}. An edit that passes practice is frozen, and the operator scores it on \emph{held-out seeds} never used during editing. The two policies alternate run order between the halves of the slots. We report paired differences with 95\% bootstrap intervals over the four outcome types (20,000 draws), exact McNemar tests on the fixed and broke pairs~\citep{mcnemar1947note}, and Wilson intervals for single rates~\citep{wilson1927probable}. Because paired seeds do not make episodes identical, we also rerun unedited and identically edited policies as noise controls.

\section{Results}
\label{sec:results}
We first report the distilled edit on its own task, its size, and its transfer (Sections~\ref{sec:main}--\ref{sec:transfer}). We then ask where in the network the same distillation works (Section~\ref{sec:where}) and whether the released tool reproduces it (Section~\ref{sec:tool}). The remaining subsections cover the blind editors (Section~\ref{sec:blind}), edits at the other failure levels (Section~\ref{sec:levels}), and the noise floor of the comparisons (Section~\ref{sec:noise}).

\subsection{The distilled edit on held-out seeds}
\label{sec:main}
\begin{figure}[!t]
\centering
\includegraphics[width=\linewidth]{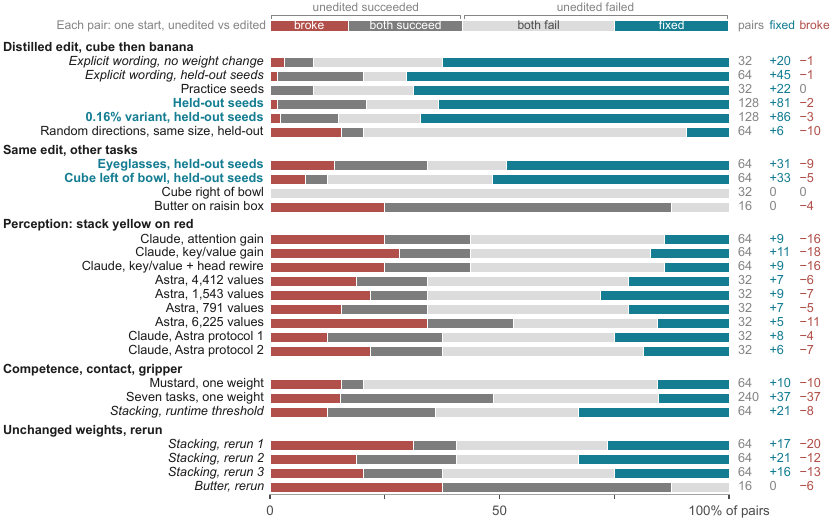}
\caption{Paired outcomes. Each bar splits a comparison's paired starts into the four outcome types, with the number of pairs and of fixed and broke pairs at right. Held-out comparisons are in bold teal, and italic rows change no weights. On its own task the distilled edit flips pairs almost only one way. Failed edits and unchanged-weight reruns flip many pairs both ways. Appendix~\ref{app:comparisons} shows every comparison.}
\label{fig:outcomes}
\end{figure}

Closed-form instruction distillation fixes the misreading on seeds never used during editing. On 128 held-out seeds, success rises from 27/128 (21.1\%) to 106/128 (82.8\%), with 81 episodes fixed and 2 broken (Fig.~\ref{fig:outcomes}). The paired difference is +61.7 percentage points (95\% interval 52.3--70.3; exact McNemar $p=7.2\times10^{-22}$), and the Wilson intervals do not overlap (75.3--88.4\% against 14.9--29.0\%). The gain holds in both halves of the alternating run order (Appendix~\ref{app:repro}). Successful episodes also finish sooner, in 300 rather than 412 steps on average. In the event logs, the unedited policy drops the cube during transport, outside the bowl, in 75 of the 81 fixed starts. The edited policy does so in 15 of them and recovers.

The fix comes from the edit's directions, not its size. On the first 64 of these seeds, a random-direction update with the same layers, rank, and singular values scored 9/64. That is in line with the unedited policy's 13/64 and 9/64 in two runs ($p=0.45$ against the first). The distilled edit scored 50/64 on the same seeds. Given the explicit wording, the unedited policy scored 57/64 ($p=0.12$ against the edit), so the edit recovers most of the wording's effect (Appendix~\ref{app:construction}).

\subsection{A smaller edit}
\label{sec:small}
The edit can be made much smaller and keep most of its effect. Restricting the keys to the 32 image cells in layers 8--16 gives an update of rank at most 32 per layer. Stored as low-rank factors, it has 5.3 million parameters, or 0.16\% of the parameter count. On the same held-out seeds this variant scores 102/128 against 19/128 (86 fixed, 3 broken; exact $p=3.8\times10^{-22}$). On the first 64 of them it scores 54/64, against 57/64 for the explicit wording (7 fixed, 10 broken; $p=0.63$). It does not keep all of the effect, however. Head to head on 64 new held-out seeds, it scored 41/64 against the full edit's 55/64 ($p=0.007$; Section~\ref{sec:where}).

\subsection{Transfer to other tasks}
\label{sec:transfer}
The same edit transfers to two other tasks and harms none detectably. Its keys sit at positions every task has, so we ran it unchanged on all 15 other surveyed tasks (Appendix~\ref{app:sidefx}). On held-out seeds, picking up the eyeglasses rises from 22/64 to 44/64 (31 fixed, 9 broken; +34.4 points, 95\% interval 17.2--51.6; $p=6.8\times10^{-4}$). Putting the cube left of the bowl rises from 8/64 to 36/64 (33 fixed, 5 broken; +43.8 points, 28.1--59.4; $p=4.3\times10^{-6}$). Putting the cube right of the bowl stays at 0/32, so the edit is not a general switch for spatial language. No difference on the remaining 13 tasks is significant (all exact $p\geq0.12$). The largest apparent loss is butter on the raisin box, at 10/16 against 14/16. It is smaller than the unedited policy's own drop to 8/16 when rerun on the same seeds.

\subsection{Where to edit}
\label{sec:where}
\begin{figure}[!t]
\centering
\includegraphics[width=\linewidth]{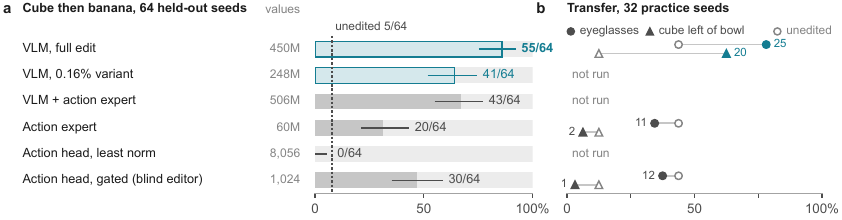}
\caption{Where to edit. (a) Held-out success of each placement of the distillation on 64 shared seeds, with Wilson 95\% intervals. The dashed line marks the shared unedited run (5/64), and Values counts the stored scalar values changed. (b) Success of the placements that were run on the transfer tasks, on 32 paired practice seeds each, unedited (open markers) and edited (filled). Only the VLM edit transfers.}
\label{fig:where}
\end{figure}

To test whether the location of the edit matters, we built the same distillation into other parts of the policy, from the same 6 recorded premature releases and the same explicit wording. We scored every variant against one shared unedited run on 64 new held-out seeds (94016--94079). We then ran them on 32 paired practice seeds of the two transfer tasks and of a retention task, cube into the bowl (Table~\ref{tab:where}, Fig.~\ref{fig:where}). The action-expert edit solves the same ridge problem for the action expert's MLP down-projections, through the time-dependent gate that scales each MLP output. The action-head edit is the least-norm change to the action output projection. The gated head edit is the blind editor's accepted edit (Section~\ref{sec:blind}), which uses negative examples that distillation does not. Construction details are in Appendix~\ref{app:where}.

\begin{table}[!t]
\centering\small
\caption{The same distillation placed in different parts of the policy. Held-out: successes of 64 (unedited 5/64), fixed/broke, and exact McNemar $p$ against the unedited run. Eyeglasses and Cube left: successes of 32 paired practice seeds (unedited: eyeglasses 14, cube left of bowl 4). Releases: recorded premature releases (of 21) left offline. The VLM edit scores highest on every task.}
\label{tab:where}
\setlength{\tabcolsep}{4pt}
\begin{tabular}{@{}lrrlcc@{}}
\toprule
Edit location & Values & Releases & Held-out & Eyeglasses & Cube left\\
\midrule
VLM (full edit) & 450M & 0 & \textbf{55} (50/0), $p=1.8\times10^{-15}$ & \textbf{25} & \textbf{20}\\
VLM, 0.16\% variant & 248M & 2 & 41 (38/2), $p=1.5\times10^{-9}$ & --- & ---\\
VLM + action expert & 506M & 0 & 43 (41/3), $p=1.6\times10^{-9}$ & --- & ---\\
Action expert & 60M & 1 & 20 (20/5), $p=0.004$ & 11 & 2\\
Action head, least norm & 8{,}056 & 2 & 0 (0/5), $p=0.06$ & --- & ---\\
Action head, gated (blind editor) & 1{,}024 & 0 & 30 (28/3), $p=4.6\times10^{-6}$ & 12 & 1\\
\bottomrule
\end{tabular}
\end{table}

The fix lives in the VLM. The VLM edit scores 55/64 against 5/64, with 50 starts fixed and none broken. Every other placement scores below it on the target task (all $p\leq0.017$). It is also the only placement that transfers, with eyeglasses at 25 against 14 of 32 ($p=0.019$) and cube left of the bowl at 20 against 4 ($p=3.1\times10^{-5}$). On the retention task, every placement run there stayed at 31--32 of 32. The VLM edit changes each layer's weights by a median 1.7\% of their norm. At the distilled decisions, it reproduces 104--114\% of the explicit wording's change in every action-expert layer. The action expert thus computes what it computes under the explicit wording without being touched.

Copied further downstream, the same fix needs a larger change and spills into other decisions. In the action expert, the minimum-norm update is about as large as the weights themselves (median 0.99 of their norm), because the gate scales the MLP output down. Applying Eq.~\eqref{eq:edit} while ignoring the gate left all 21 releases. Solved through the gate, the update fixes 20 starts and breaks 5. Its first action chunk closes the gripper in 52\% of episodes, before the arm has reached anything (Table~\ref{tab:where-stages}). The least-norm head edit pushes the gripper toward closing almost equally at the premature release and on the approach to the banana. It closes the gripper in the first chunk of 94\% of episodes and fixes nothing. A gate is what makes a head edit work. The blind editor's gated edit scores 30/64, about half of the VLM edit's gain, and transfers to neither task. Adding the action-expert update to the VLM edit matched the explicit wording best offline. In closed loop, however, it did worse than the VLM edit alone (43 against 55, $p=0.017$), mainly in the banana step. The fix belongs where the known-good input differs from the failing one. Here, that is how the VLM reads the instruction.

\subsection{The method as a tool}
\label{sec:tool}
We release the procedure as a command, \texttt{weightsmith distill}, that runs its steps end to end against a project. It selects failure decisions from the unedited policy's practice episodes by a rule and replays them under the original and the explicit wording. It then builds the edit of Eq.~\eqref{eq:edit} and checks it offline on decisions not used to build it. It never starts a closed-loop evaluation itself. From the stored capture, its build step reproduces both published edits bit for bit, with the same fingerprints (Appendix~\ref{app:repro}).

To test the whole pipeline, we ran it on a fresh cube-then-banana project with its own practice episodes and with the recipe of the 0.16\% variant (image cells, layers 8--16). The decisions were chosen by the tool's early-release rule rather than by hand. The rule needed one change, found with the tool's offline check, to build only from releases during the first carry. The operator then scored the frozen edit like every submission, on 64 held-out seeds never used before (95008--95071, run order alternating by half). Success rose from 8/64 to 36/64, with 33 starts fixed and 5 broken (exact McNemar $p=4.3\times10^{-6}$; Wilson intervals 6.5--22.8\% and 44.1--67.7\%). The gain is smaller than that of the hand-built edits on their seeds, but the decisions, the capture, and the edit all came from the tool. The tool has been run on this one task only.

\subsection{Seven blind editors}
\label{sec:blind}
To test whether an autonomous editor can find such an edit without our help, we gave seven fresh Claude Opus editors one RoboLab task each, one per failure kind (Appendix~\ref{app:study}). They had the same brief, tools, and budget, and each was blind to the others, to this paper, and to the web. Each editor diagnosed its task with activation captures, in which instruction overrides were allowed for diagnosis only. It then built at most two closed-form candidates within 96 practice episodes and submitted one candidate or the unedited policy. The operator scored every submission on 64 held-out slots that no editor saw (Table~\ref{tab:study}).

\begin{table}[!t]
\centering\small
\caption{Seven blind Claude Opus editors, one task per failure kind. Development: unedited $\to$ candidate successes on 32 paired slots (70000--70031), with fixed/broke. Held-out: the operator's score of the submission on 64 paired slots (91008--91071), with the exact McNemar $p$. Values: entries changed by the submitted edit. Only the cube-then-banana submission passed. $^*$Evaluated but not submitted. $^\dagger$After the study, with more time and the same two-candidate budget.}
\label{tab:study}
\setlength{\tabcolsep}{2.8pt}
\begin{tabular}{@{}lllrllr@{}}
\toprule
Task & Failure kind & Submission & Values & Development & Held-out & $p$\\
\midrule
Cube then banana & Two-step instruction & edit & 1,024 & 4 $\to$ 10 (8/2) & \textbf{11 $\to$ 27} (22/6) & 0.004\\
Pick up eyeglasses & Thin-object pickup & edit & 3,072 & 12 $\to$ 18 (9/3) & 18 $\to$ 27 (18/9) & 0.12\\
Mustard into left bin & Grasp and carry & edit & 4,094 & 6 $\to$ 9 (7/4) & 8 $\to$ 11 (9/6) & 0.61\\
Cube left of bowl & Spatial relation & unedited & --- & 5 $\to$ 3 (3/5)$^*$ & --- & \\
Stack three cubes & Assembly & unedited & --- & 16 (no candidate) & --- & \\
Stack yellow on red & Lookalike perception & unedited & --- & 13 (no candidate) & --- & \\
Pick up flat box & Flat-object pickup & unedited & --- & 0 $\to$ 0 (0/0)$^*$ & --- & \\
\midrule
Stack three cubes$^\dagger$ & Assembly & edit & 3,072 & 16 $\to$ 17 (6/5) & 35 $\to$ 39 (17/13) & 0.585\\
\bottomrule
\end{tabular}
\end{table}

One of seven submissions passed. On cube then banana, the editor independently traced the release to the wording. Replaying 16 premature releases under ``Put the cube in the bowl'' turned all 16 into holds. It then wrote an edit different from ours, a rank-one change of 1,024 values to the gripper column of the action head, gated to premature-release states. The edit took held-out success from 11/64 to 27/64 (22 fixed, 6 broken; $p=0.004$, or 0.026 after Holm correction over the seven tasks). Edits for the eyeglasses and the mustard stayed within noise, and four editors submitted the unedited policy. On lookalike perception, no rewording moved the arm, so there was nothing to distil.

A second round on stacking shows what a correct decision edit cannot do. Given more time after the study, with the same budget, the stacking editor built the gate it had proposed. The gate is one repurposed unit in the action expert's last MLP (3,072 values). It holds the gripper open when the empty gripper would close above the two-cube stack. The edit removed the decision it targeted. On the 32 development episodes, closes on empty air at the stack fell from 73 to 5 replans. Success did not move, at 17/32 against 16/32 on development seeds and 39/64 against 35/64 on held-out seeds (17 fixed, 13 broken; $p=0.585$). The remaining failure is placing the third cube, a matter of motor precision that the edit does not touch. The edit fixed a wrong decision, but it could not supply the missing placement skill.

\subsection{Edits at the other failure levels}
\label{sec:levels}
Edits at the other failure levels did not help (Fig.~\ref{fig:outcomes}). In perception, the unedited policy stacks the yellow block on the red block in 28 of 64 practice episodes. In 48 of 63, the arm goes first for a nearby Rubik's cube with a yellow face. The VLM separates the two objects ($d'\approx10$--$20$), but the action expert barely uses the distinction ($d'\approx2$--$3$, against a null of about 1.7). Three Claude edits amplified this signal through attention or key and value weights. Each scored 21/64 against 28/64 and also suppressed a useful habit of backing away from the cube after touching it. Four GPT-6 Astra edits of 791--6,225 values stayed within noise, and none passed the companion study's gate. A fresh Claude Opus editor, working under that study's exact protocol, independently diagnosed the same lookalike confusion. Its two proposals scored 16/32 and 11/32 against 12/32. Neither passed the gate, and the editor submitted the unedited policy (Appendix~\ref{app:other}).

The other levels gave the same answer. The flat green box is never picked up (0/64), because its long side exceeds the gripper opening at the default wrist angle. The best of ten edits raised the share of episodes reaching a graspable wrist angle from 9/16 to 16/16 but produced no success. The mustard bottle slips from a cap grasp mid-carry, and no recorded decision variable predicted which grasps would hold ($|\rho|\leq0.26$). The companion study's lower gripper threshold~\citep{abbas2026wspstudy}, built into the network as one bias weight, scored 117/240 against 117/240 across seven tasks, with 37 episodes fixed and 37 broken.

\subsection{The noise floor}
\label{sec:noise}
\begin{figure}[!t]
\centering
\includegraphics[width=\linewidth]{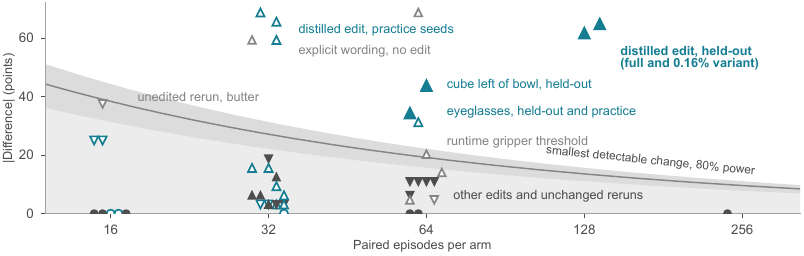}
\caption{Which differences a paired comparison can detect. Each marker places a comparison by its paired episodes and the size of its difference, and points up when the edit did better. Below the curve, a difference is smaller than the change detectable with 80\% power ($\alpha=0.05$, 30\% of pairs discordant, band 20--40\%). Teal marks the distilled edit (filled when held-out), gray no weight change, and dark the other edits.}
\label{fig:detect}
\end{figure}

Paired seeds still leave large run-to-run variation. The unedited policy scored 14/16 and then 8/16 on the same 16 butter seeds. Two runs of one edited policy agreed on only 46 of 64 paired episodes. A near-neutral one-weight edit changed 70 of 192 paired outcomes on tasks with room to move. With about 30\% of pairs discordant, a paired comparison with 80\% power at $\alpha=0.05$ detects changes of about 27 points with 32 pairs, 19 with 64, 14 with 128, and 10 with 256 (Fig.~\ref{fig:detect}). A 16-episode screen is therefore a filter, not evidence. The unedited policy's own success on cube then banana ranged from 5/64 (7.8\%) to 27/128 (21.1\%) across the held-out runs above. Because every comparison is paired, this moves absolute rates, not the differences.

\section{Why Some Failures Can Be Edited}
\label{sec:why}
The image tokens carry the fix. Three simpler edits failed offline (Fig.~\ref{fig:construction}b). Steering image tokens along a cube-versus-banana axis that tracks the wording flipped none of the 21 releases in either of two attempts. That axis was correlated with the outcome but did not cause it~\citep{turner2023steering}. Copying the mean shift at the state tokens flipped none, and per-token shifts at the 40 prompt and state positions left 8 to 15. Adding the 32 image cells removed all 21. Image cells alone suffice, whereas prompt and state positions alone leave 13 (Appendix~\ref{app:construction}). Placed in the action expert or head instead, the same distillation helps less or not at all and does not transfer (Section~\ref{sec:where}). The explicit wording moves the image tokens by only 2--4\% of their norm, yet the action expert places 13--44\% of its attention on them. Because prefix attention is bidirectional, image tokens attend to the instruction. Our working hypothesis is that the instruction is read into the image representation that the action expert consults when it decides whether to keep holding.

Distillation re-routes a computation the network already performs. It did not create one that is missing. The success had three ingredients. The policy demonstrably has the skill, the failure traces to one computation, and there is a known-good internal state to copy. Each failed level lacked one of these. The flat box lacks the skill, and the mustard grasp has no wrong decision to correct. In perception, no edit we found routed the VLM's correct signal to the action without breaking a useful behavior. This agrees with the strongest outside reports, which edited checkpoints already fine-tuned on each task~\citep{guo2026robotweights} or built updates from differences between trained models~\citep{gupta2026robotweights,ilharco2023arithmetic}. This suggests a practical rule. Edit when the network can be shown to do the right thing under some input, and train when it cannot.

The blind editors support the rule from both sides (Section~\ref{sec:blind}). The one editor whose task had a known-good wording found it and passed held-out scoring. Where no rewording moved the arm, there was nothing to distil. The stacking gate removed the wrong decision it targeted without raising success, because what remained was a missing skill. The edits fixed decisions, but none supplied a skill.

These conclusions have limits. The full edit is low-rank but not sparse. It changes 13.4\% of the policy's stored values, and the 0.16\% variant still changes 7.4\% once applied. The distilled edit was built by the operator session under the editor rules. The released tool rebuilds such an edit end to end, but it has run on one task, and it needs the known-good wording as input. Discovery needs a known-good instruction. The method turns such a prompt fix into a weight fix, which helps when the instruction cannot be changed but not when no input succeeds. The comparison of edit locations covers one task and one wording. Transfer was not predicted in advance (left of the bowl but not right), so deployment would need broad regression checks. All results come from one zero-shot checkpoint in simulation, with 16 to 128 pairs per comparison. The failure levels were assigned without an independent rater.

\section{Concluding Remarks}
Closed-form instruction distillation fixed a misread instruction in a zero-shot generalist VLA without gradients or a change to the instruction. A random update of the same size did not. The fix lives in the VLM, where the instruction is read. Copied into the action expert or the action head, it helped less or not at all and did not transfer. The released tool reproduced the fix end to end on new held-out seeds, and a blind editor repeated the diagnosis. Edits aimed at the other failure levels did not help, and an edit that removed a wrong decision in stacking left success unchanged. The practical rule is to find an input under which the policy succeeds, distil its internal state, and confirm the gain on held-out seeds.

Future work should test task-fine-tuned checkpoints, smaller edits, other tasks for the released tool, and a physical robot.

\clearpage
\bibliography{refs}
\clearpage
\appendix

\setlength{\parskip}{5.5pt}
\section{Unedited-policy survey and failure levels}
\label{app:survey}
Table~\ref{tab:survey} lists the 16 RoboLab tasks with the unedited policy's success in 16 episodes and its dominant failure classes. Episodes last 450 or 900 control steps at 15\,Hz, depending on the task. Failure classes come from RoboLab events and recorded gripper commands. The ``wrong object first'' class relies on a RoboLab event that the companion study's audit found to over-count~\citep{abbas2026wspstudy}, so those counts are upper bounds. Levels for the five case-study tasks rest on editor diagnoses from kinematics, camera frames, and activation captures. No independent rater checked them. Table~\ref{tab:levels} summarizes each level and what weight editing did at it.

\begin{table}[htbp]
\centering\small
\caption{Unedited-policy success in 16 episodes per task, with the dominant failure classes among the failed episodes.}
\label{tab:survey}
\begin{tabular}{@{}p{.36\linewidth}rp{.44\linewidth}@{}}
\toprule
Task & Success & Dominant failure classes\\
\midrule
\input{generated/survey_table.tex} 
\bottomrule
\end{tabular}
\end{table}

\begin{table}[htbp]
\centering\small
\caption{Failure levels and what weight editing did at each. Counts are edited versus unedited on paired seeds unless stated. Editing helped only at the language-binding level.}
\label{tab:levels}
\begin{tabular}{@{}>{\raggedright\arraybackslash}p{.16\linewidth}>{\raggedright\arraybackslash}p{.26\linewidth}>{\raggedright\arraybackslash}p{.15\linewidth}>{\raggedright\arraybackslash}p{.32\linewidth}@{}}
\toprule
Level & What goes wrong & Example & Weight editing\\
\midrule
Language binding & The wrong object counts as the current target & Cube then banana & Held-out 106/128 vs 27/128; transfers to eyeglasses and cube left of bowl\\
Perception & Goes for a similar-looking object & Stack yellow on red & 3 Claude edits at $-$10.9 points; 4 Astra and 2 same-protocol Claude proposals, none accepted\\
Missing competence & Never produces a working motion & Pick up flat box & 10 edits change the motion, 0 successes\\
Contact physics & Decisions look right, the grasp slips & Mustard into bin & No decision-level lever; edits stopped offline; one weight 13/64 vs 13/64\\
Gripper commitment & Open and close decisions near the threshold & Seven tasks & One weight, 117/240 vs 117/240\\
Motor decision, skill present & A competent skill stalls at one step & None among our 16 tasks & Not tested; others report gains on task-fine-tuned checkpoints\\
\bottomrule
\end{tabular}
\end{table}

\FloatBarrier
\section{All paired comparisons and the noise floor}
\label{app:comparisons}
\begin{figure}[p]
\centering
\includegraphics[width=\linewidth]{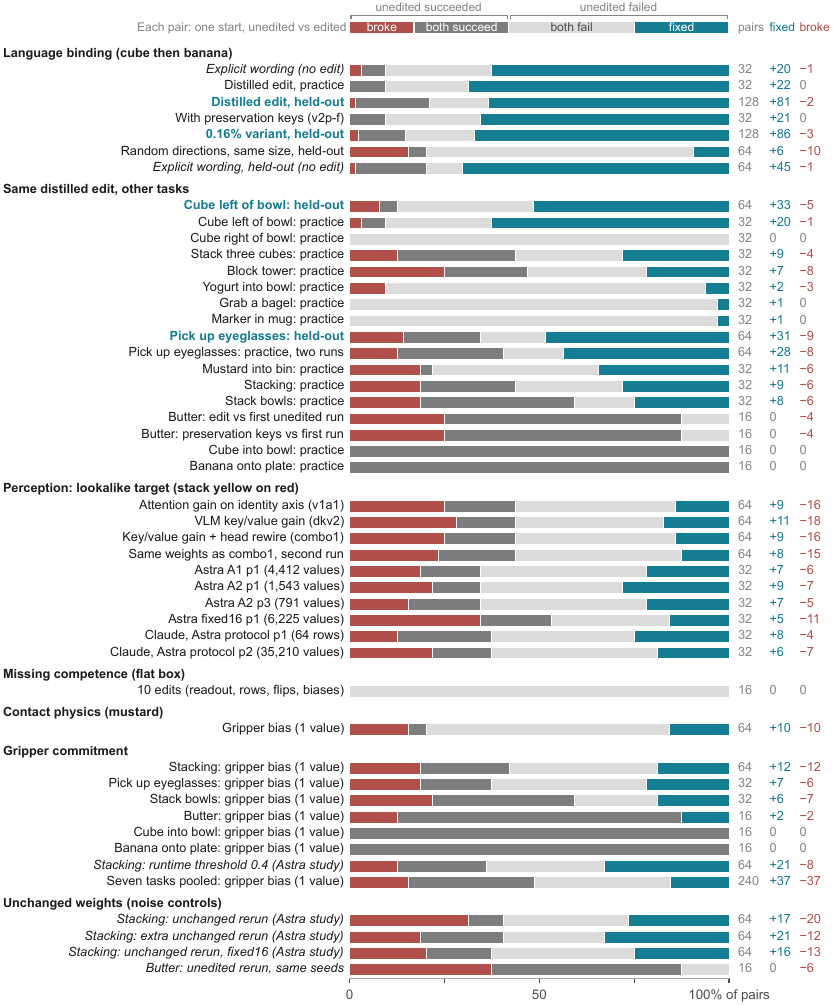}
\caption{Paired outcomes of every comparison in Table~\ref{tab:all}, drawn as in Fig.~\ref{fig:outcomes}. Held-out comparisons are in bold teal. The flat-box row stands for each of the ten edits, all 0/16 against an unedited 0/64.}
\label{fig:outcomes-full}
\end{figure}

Table~\ref{tab:all} lists every closed-loop comparison in \texttt{study/results/opus/comparisons.json}, and Fig.~\ref{fig:outcomes-full} shows the paired outcomes of each. Each difference is edited minus unedited on paired seeds, in percentage points, with a 95\% interval from 20,000 multinomial draws over the four paired outcome types (both succeed, fixed, broken, both fail). The pooled gripper row is resampled within each task. The $p$-values are exact two-sided McNemar tests on the fixed and broken pairs. The companion study's rows use its own runtime and are not pooled with ours. Its unchanged-weight rerun labelled ``extra'' shares the unedited reference of the first rerun, so those two comparisons are correlated. Two runs of one edited policy (\texttt{combo1}) on the same seeds agreed on only 46 of 64 paired episodes. The random-direction control and the held-out explicit-wording arm are paired with the first unedited run on their seeds. The controlled study of Appendix~\ref{app:study} is reported separately, in Table~\ref{tab:study}. Rows that begin with Astra are GPT-6 Astra edits, and rows marked Astra study are the companion study's operator controls. All other edits, including the two under the Astra protocol, are Claude Opus edits.

{\fontsize{8}{9.6}\selectfont
\setlength{\tabcolsep}{2.6pt}
\setlength{\LTcapwidth}{\linewidth}
\begin{longtable}{@{}>{\raggedright\arraybackslash}p{.27\linewidth}rlcll@{}}
\caption{Every paired closed-loop comparison. Values: stored scalar values changed by the edit (--- for none or not recorded). Counts: edited versus unedited successes on the same seeds. Difference in percentage points.}
\label{tab:all}\\
\toprule
Comparison & Values & Counts & Fixed / broke & Difference [95\% CI] & $p$\\
\midrule
\endfirsthead
\toprule
Comparison & Values & Counts & Fixed / broke & Difference [95\% CI] & $p$\\
\midrule
\endhead
\input{generated/comparisons_table.tex} 
\bottomrule
\end{longtable}
}

\paragraph{Noise floor.} Unchanged weights do not give unchanged outcomes on paired seeds. The share of paired episodes whose outcome differs is 28\% for two runs of one edited policy, 36\% for a near-neutral one-weight edit on the four tasks with room to move, 38\% for the unedited butter rerun, and 45--58\% for the companion study's unchanged-weight reruns (Fig.~\ref{fig:outcomes-full}). The companion study's audit found different first camera images and actions in paired episodes with equal seeds and robot state, without isolating the cause~\citep{abbas2026wspstudy}. One candidate is nondeterministic GPU simulation, amplified by open-loop action chunks. Table~\ref{tab:mde} gives the smallest paired difference a single comparison detects with 80\% power at a two-sided $\alpha=0.05$, using the normal approximation $(z_{0.975}+z_{0.8})\sqrt{q/n}$ with a discordant fraction $q=0.3$. Fig.~\ref{fig:detect} places every observed effect against that bound.

\begin{table}[htbp]
\centering\small
\caption{Smallest detectable paired difference with about 30\% discordant pairs (80\% power, $\alpha=0.05$).}
\label{tab:mde}
\begin{tabular}{@{}rr@{}}
\toprule
Paired episodes per arm & Smallest detectable change\\
\midrule
32 & about 27 points\\
64 & about 19 points\\
128 & about 14 points\\
256 & about 10 points\\
\bottomrule
\end{tabular}
\end{table}

\FloatBarrier
\section{Edit construction, failed attempts, and ablations}
\label{app:construction}
\paragraph{Capture and tensors.} The edit uses one activation capture of 6 premature-release decisions, each replayed under 8 instructions, with VLM residuals and MLP activations at every layer. It modifies the MLP down-projections \texttt{language.blocks.00--16.mlp.linear}, each mapping 16,384 hidden features to the 2,048-dimensional residual stream. Keys are mean activations under the original instruction. Values are the explicit instruction's per-layer residual increments minus the original's. Each layer receives Eq.~\eqref{eq:edit} with $\lambda=10^{-3}\,\mathrm{tr}(KK^{\top})/n$ for $n$ keys and unit strength. After rounding to bfloat16, the full edit changes 450,468,837 stored values (13.4\%) with rank at most 72 per layer.

\paragraph{Attempts that failed offline.} Offline replay on recorded decisions stopped each of these attempts before any rollout (Fig.~\ref{fig:construction}b). Two rank-one pushes along a cube-versus-banana ``target'' axis at the image tokens left 21 of 21 releases. Copying the mean explicit-minus-original shift at the state tokens, rank one per layer, left 6 of 6 releases on a six-decision check. Copying the per-position shift at the 40 prompt and state positions with Eq.~\eqref{eq:edit} left 15, 13, and 8 of 21 releases at strengths 0.5, 1, and 2 (candidates \texttt{v2m-h}, \texttt{v2m-f}, \texttt{v2m-2}). Adding text positions left 13. Adding the 32 image cells gave the submitted edit, with 0 of 21 releases and unchanged carries and phase-two decisions.

\paragraph{Ablations and the 0.16\% variant.} Table~\ref{tab:ablation} lists the offline ablations shown in Fig.~\ref{fig:construction}b, with their layers, keys, and sizes. Applied to the weights, every variant still changes 248--450 million stored values (7--13\%), because each low-rank update is dense in the weight matrix. The smallest variant (image cells, layers 8--16, rank at most 32) changes 248,057,499 values. In closed loop it scores 25/32 against 3/32 on the practice seeds (22 fixed, 0 broken). On held-out seeds 90000--90127 it scores 102/128 against 19/128 (86 fixed, 3 broken; exact $p=3.8\times10^{-22}$), with Wilson intervals of 71.9--85.7\% and 9.7--22.0\%. Successful episodes take 333 rather than 396 steps on average.

\paragraph{Random-direction control.} The control keeps the size of the distilled update and randomizes its directions. For each of layers 0--16 we recomputed the distilled update in float64 from the same capture. It reproduces the stored edit exactly after bf16 rounding. We kept its 72 singular values and replaced its left and right singular vectors with random orthonormal bases (NumPy \texttt{default\_rng(0)}). The control therefore matches the edit's layers, rank, singular values, and Frobenius norm in every layer. It changes 502,291,925 stored values (15.0\%), against the edit's 450,468,837 (13.4\%). It was frozen and fingerprinted like a submission, and the operator ran it on held-out seeds 90000--90063. It scored 9/64, against 13/64 and 9/64 for the unedited policy's two runs on those seeds (6 fixed and 10 broken, exact $p=0.45$; 4 and 4, $p=1$). Against the distilled edit's 50/64, it has 0 fixed and 41 broken pairs. In the same run, the explicit wording given to the unedited policy scored 57/64 (45 fixed and 1 broken against the first unedited run, $p=1.3\times10^{-12}$). Against the explicit wording, the distilled edit has 4 fixed and 11 broken pairs ($p=0.12$), and the 0.16\% variant 7 and 10 ($p=0.63$).

\begin{table}[htbp]
\centering\small
\caption{Offline ablations of Eq.~\eqref{eq:edit} on the 21 recorded premature releases. Every variant keeps all 9 correct carries and 19 phase-two decisions. Factored size is rank $\times$ (16{,}384 + 2{,}048) $\times$ layers, with its share of the parameter count. Held-out: edited versus unedited successes of 128.}
\label{tab:ablation}
\setlength{\tabcolsep}{5pt}
\begin{tabular}{@{}llrrrr@{}}
\toprule
Variant & Layers & Keys per layer & Factored size & Releases (of 21) & Held-out\\
\midrule
Full edit & 0--16 & 72 & 22.6M (0.67\%) & 0 & 106 vs 27\\
Image cells only & 0--16 & 32 & 10.0M (0.30\%) & 0 & ---\\
Layers 8--16 only & 8--16 & 72 & 11.9M (0.36\%) & 0 & ---\\
Half strength & 0--16 & 72 & 22.6M (0.67\%) & 2 & ---\\
Image cells, layers 8--16 & 8--16 & 32 & 5.3M (0.16\%) & 2 & 102 vs 19\\
Prompt and state positions only & 0--16 & 40 & 12.5M (0.37\%) & 13 & ---\\
\bottomrule
\end{tabular}
\end{table}

\paragraph{Diagnosis details.} The destination words' attention to the adjacent noun versus the first noun is 0.69 versus 0.02 in VLM layer 1, head 1. The explicit wording changes the image tokens by 2--4\% of their norm, while the action expert places 13--44\% of its attention on image tokens and 33--58\% on the instruction and state tokens. On recorded decisions, the edited policy's actions land 4.5\,cm from the explicit wording's actions at the end of the chunk, against 8.1\,cm for the unedited policy. Replayed again for the where-to-edit study (Appendix~\ref{app:where}), on the 21 premature releases alone, the distances are 4.7 and 8.7\,cm.

\FloatBarrier
\section{Edits at the other failure levels}
\label{app:other}
\paragraph{Perception: stack yellow on red.} The Claude cohort built three edits that amplify the VLM's yellow-versus-cube distinction in the action pathway. They are an attention gain on an identity axis (30.4 million values), a VLM key and value gain (12.0 million), and that gain combined with an attention-head rewire (12.2 million). Each scored 21/64 against 28/64. A second run of the combined edit with identical weights scored 21/64 again, with different fixed and broken episodes (8 and 15 instead of 9 and 16). Several further stacking candidates were built but never evaluated in closed loop. The companion study's four Astra edits changed 4,412, 1,543, 791, and 6,225 values in selected action-expert MLP rows and scored 12/32, 13/32, and 13/32 against 11/32, and 11/32 against 17/32 in a separate runtime~\citep{abbas2026wspstudy}.

\paragraph{Claude Opus under the companion protocol.} We gave a fresh Claude Opus editor the companion study's brief for the stacking task. It received the same 32 development episodes, at most three candidate evaluations on those slots, 96 activation items, and 7,200 seconds. Edits were limited to the action expert's MLP output rows, at most 128 rows across two layers, through typed operations (row scaling, adding a direction to rows, projecting a direction out of rows), with no fitted regression or search. The fixed gate accepts a candidate only with strictly more successes than the unedited 12/32 and at most two newly broken pairs. The editor found that the main failure is confusion with yellow-looking distractors, decided by noise in the first action chunks. In 13 of 32 unedited episodes the first grasp was the Rubik's cube, and 3 of those 13 succeeded, against 9 of 17 when the yellow block came first. Its first proposal added a direction to 64 rows of the last action-expert layer. It scored 16/32 against 12/32 with 8 fixed and 4 broken, so the gate rejected it. The editor's own check found that the predicted shift of the first chunk toward the yellow block did not occur beyond run-to-run variation. A half-strength second proposal scored 11/32 (6 fixed, 7 broken). The editor declined a third proposal as ``another dose of a falsified lever, effectively a lottery draw,'' and submitted the unedited policy after 2 of 3 evaluations, 64 of 96 activation items, and about 7,000 of 7,200 seconds, for about \$20 of API use.

\paragraph{Missing competence: pick up the flat box.} The unedited policy is 0/64. The box's long side is wider than the gripper opening, so at the default wrist angle the fingers land on top of it. When the wrist does rotate, the close command flip-flops. The ten edits tried were readout surgery, neuron flips, a ``stay closed'' latch, and wrist and gripper biases. Nine scored 0/16 and one 0/32. The best raised the share of episodes reaching a graspable wrist angle from 9/16 to 16/16 in closed loop, without a success.

\begin{figure}[htbp]
\centering
\includegraphics[width=\linewidth]{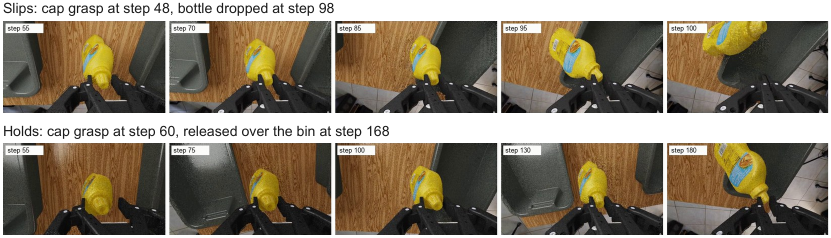}
\caption{Wrist-camera frames of the unedited policy on mustard into bin. Both rows grip the cap. The top grasp slips and the bottom one holds, and no recorded decision variable separated them.}
\label{fig:mustard}
\end{figure}

\paragraph{Contact physics: put the mustard in the bin.} The unedited policy succeeds in 16 of 96 episodes. It grabs the bottle by its narrow cap and keeps the gripper commanded closed, and in most episodes the bottle slips mid-carry (Fig.~\ref{fig:mustard}). Close height, wrist angle, timing, and carry speed did not predict which grasps hold ($|\rho|\leq0.26$). Two neuron-flip edits in the style of Gupta et al.~\citep{gupta2026robotweights} were built from 192 recorded grasps and changed 131,072 and 1,048,576 values. They either left the actions nearly unchanged (0.0005\,rad mean joint change and 0 of 96 flipped decisions) or pushed them the wrong way, and both were stopped offline. The one-weight gripper edit scored 13/64 against 13/64.

\paragraph{Gripper commitment.} The companion study's largest effect came from a runtime change that binarizes the gripper at 0.4 instead of 0.5. It scored 36/64 against 23/64 on stacking, while that study's own unchanged-weight repeat also moved from 26 to 35. We built the same shift into the network as one scalar, the gripper output bias. On fixed recorded inputs it raised the mean gripper command by 0.109 and the number of closed rows from 674 to 781 of 1,440. In closed loop it was exactly neutral (Table~\ref{tab:gripper}).

\begin{table}[htbp]
\centering\small
\caption{One-weight gripper edit, edited versus unedited on paired seeds. In total, fixed and broken pairs cancel.}
\label{tab:gripper}
\begin{tabular}{@{}lrrrr@{}}
\toprule
Task & Pairs & Unedited & Edited & Fixed / broke\\
\midrule
\input{generated/gripper_table.tex} 
\bottomrule
\end{tabular}
\end{table}

\FloatBarrier
\section{Side effects and transfer}
\label{app:sidefx}
\paragraph{Transfer map.} Table~\ref{tab:transfer} lists the distilled edit on all 15 other surveyed tasks. The first seven other tasks were compared with an earlier unedited run on the same slots. The transfer map of eight more tasks ran the unedited and edited policies back to back on the same 32 slots. Held-out runs used slots 90000--90063, with the run order alternating by half. Eyeglasses failures are often grasps that are never lifted. We hypothesize that the edit strengthens a generic signal that the held object is the one to carry, so the policy keeps hold and persists. Successful held-out eyeglasses episodes do take longer with the edit (224 versus 158 steps on average). The left-of-bowl gain without a right-of-bowl gain shows that the edit is not a general switch for spatial language. Our working hypothesis is that it strengthens how the network ties a destination phrase to the carried object, and that something else fails first in the right-of-bowl task. Neither hypothesis has been tested.

\begin{table}[htbp]
\centering\fontsize{8}{9.6}\selectfont
\setlength{\tabcolsep}{3.2pt}
\caption{The distilled edit on the other surveyed tasks, edited versus unedited on paired seeds. Difference in percentage points with its paired 95\% bootstrap interval, and $p$ from the exact McNemar test. The last row is the unedited butter rerun. Only the eyeglasses and cube-left-of-bowl gains are significant.}
\label{tab:transfer}
\begin{tabular}{@{}lrrrcll@{}}
\toprule
Task & Pairs & Unedited & Edited & Fixed / broke & Difference [95\% CI] & $p$\\
\midrule
\input{generated/transfer_table.tex} 
\bottomrule
\end{tabular}
\end{table}

\paragraph{The butter identity rerun.} The butter comparison cannot separate an edit effect from the policy's own run-to-run variation. Butter on the raisin box first looked like a loss, 10/16 against 14/16 with 0 fixed and 4 broken. Rerun on the same 16 seeds with the same weights, the unedited policy itself scored 8/16 (0 fixed and 6 broken against its first run; exact $p=0.031$). The edit's paired bootstrap interval against the first run excludes zero ($-50.0$ to $-6.2$ points; exact $p=0.12$). So does the unedited rerun's interval against the same run, with a larger drop ($-62.5$ to $-12.5$).

\paragraph{Preservation keys.} We rebuilt the edit with image-cell keys plus 160 preservation keys (32 image cells from each of five other tasks, each mapped to a zero change) in layers 0--16. Offline it left 1 of 21 releases. In closed loop it kept the fix, 24/32 against 3/32 with 21 fixed and none broken, and scored the same on butter (10/16). There was no real side effect for it to remove.

\paragraph{Evaluation lanes.} The distilled-edit runs used two evaluation lanes. Across those runs, success was 249/448 on one lane and 199/368 on the other, with no lane effect.

\FloatBarrier
\section{Controlled study with seven blind editors}
\label{app:study}
\paragraph{Design.} We gave seven fresh Claude Opus editors (Claude Code, extra-high effort) one RoboLab task each, chosen to cover one failure kind apiece (Table~\ref{tab:study} in Section~\ref{sec:blind}). Every editor received the same brief apart from the task line, and the same general hints from earlier editors, none about its own task. Each had the same tools, \texttt{owsp} and file tools inside its own directory, with no access to other projects, other editors' notes, or the web. An editor first ran the unedited policy on 32 development slots (70000--70031). It then diagnosed the failure with activation captures on recorded decisions, where instruction overrides were allowed for diagnosis only. It could build at most two candidates in closed form, with no gradients, training, fitted predictors, or search. Each candidate was evaluated once on the development slots after a written prediction. Finally, the editor submitted one candidate or the unedited policy, with a write-up. The budget was 96 practice episodes and 192 activation items. The operator scored every submitted candidate with \texttt{heldout.py} against the unedited policy on 64 held-out slots (91008--91071) that no editor saw.

\paragraph{Deviations.} The study departed from its plan in several ways. About an hour in, before any editor had evaluated a candidate, the budget was cut from three candidates to two. One identical message went to all seven editors, and the tool enforced the new limit. All seven sessions were interrupted once by the operator and resumed in fresh sessions from their own notebooks. Three base evaluations queued at that moment were lost and rerun without charge. A blocking \texttt{wait} command was added so that editors could wait for the shared GPU queue. One editor (eyeglasses) ran a single listing of its own directory outside the allowed commands. Queue waits of 25--30 minutes per job led four editors to stop after one candidate. Finally, a first held-out launch at slot 91000 stopped before any episode ran, because slots must form whole groups of 16. Scoring used slots 91008--91071. The briefs, notebooks, scripts, capture specifications and write-ups are in \texttt{study/results/ctrl/}.

\paragraph{Results.} One of seven submissions passed (Table~\ref{tab:study}). On cube then banana, held-out success rose from 11/64 to 27/64 (22 fixed, 6 broken; exact $p=0.0037$, 0.026 after Holm correction over the seven tasks). The paired difference is $+25.0$ points (95\% interval 9.4--40.6), and the gain appears in both halves of the run order (7 to 10 and 4 to 17). The Wilson intervals are 9.9--28.2\% unedited and 30.9--54.4\% edited. The eyeglasses edit gained 14.1 points (interval $-1.6$ to $+29.7$, $p=0.12$) and the mustard edit 4.7 ($-6.2$ to $+17.2$, $p=0.61$). Four editors submitted the unedited policy.

\paragraph{Findings per task.} \emph{Cube then banana.} Replaying the 16 premature releases under ``Put the cube in the bowl'' turned all 16 into holds, while ``Put the banana in the bowl'' left 5 of 6 cube holds as holds. The effect enters the action expert from layer 13 and lands almost entirely on the gripper output. The editor built a rank-one change to the gripper column of the action head (1,024 values), gated by a least-norm direction that scores 1 on premature releases and 0 on releases over the bowl, banana-stage decisions, and the approach.

\emph{Cube left of the bowl.} Prompt swaps showed that ``left'' moves the arm along a diagonal rather than the robot's $+y$ axis, leaving the cube behind the bowl. Pushing the ``left'' token toward ``front'' at layer 8 (130,764 values) corrected the carry by about a third but turned grasp closures into openings (3/32 against 5/32), so the editor submitted the unedited policy.

\emph{Stack three cubes.} Every failure builds two cubes and then closes the empty gripper on air above the stack or pulls the top cube off. A gated gripper edit failed a split-half check offline, so no candidate ran.

\emph{Stack yellow on red.} In 13 of 19 failures the arm goes first to a Rubik's cube with a yellow face. Rewording the instruction, even to name a green block, did not move the arm, so there was no known-good input to distil. A steering edit had no offline effect and was not run.

\emph{Mustard into the left bin.} After a slipping cap grasp the unedited policy is undecided between the bins (14 of 32 carries go right, 13 left). Replaying the 32 first post-grasp decisions under ``right'' sends 27 right and none left, but under ``left'' only 8 go left and 16 still go right. Two layer-0 MLP units that fire only on the token ``left'' were given rows that strengthen it (4,094 values). Carries then went left in 30 of 32 closed-loop episodes, but the grasp still slips.

\emph{Eyeglasses.} The main failure grasps one temple arm and lifts too little. The editor repurposed the least-used MLP unit of the action expert's last block as a gate (3,072 values). The gate is zero on every captured open or closing state. While holding, it adds a least-norm joint change for 8~cm of lift at the last denoising step.

\emph{Flat box.} The box is never picked up. A rank-one change to the gripper readout (1,021 values) kept closed grasps closed and lifted the box 10--19~cm in 6 episodes, but it slipped every time (0/32), so the editor submitted the unedited policy.

\emph{Shared units.} Two editors working on different tasks (cube left of the bowl, mustard) independently identified the same two layer-0 MLP units, 3097 and 8023, as detectors of the token ``left''.

\paragraph{Round 2: stacking with more time.} After the study, the stacking editor was given more time, with the same budget of two candidates, to build the gate it had proposed. In its 32 unedited development episodes, every failure completes the first two-cube stack. The empty gripper then hovers at the stack and commands a close at stack height, closing on air (73 such replans) or pulling the top cube back off. The edit repurposes unit 2437 of \texttt{action.blocks.17.mlp\_1} as a state-gated ``do not close here'' detector. It changes 3,072 values, the gate and up rows and the down row of one MLP unit. The gate direction contrasts 30 target decisions with 62 preservation decisions. The targets are an empty gripper commanding a close above the stack, or closed on air there. The preservation decisions are table-level grasps, carries, and holding a cube over the stack. The down row is the weighted least-norm write that changes the gripper output and, to first order, no joint. Offline, every preservation decision held out from the fit kept its gripper command. In closed loop on the development slots, empty-gripper closes at the stack before the third cube was picked fell from 17 to 2. Closes on air at the stack fell from 73 to 5 replans, and the third cube was reached in 27 episodes against 24. Success, however, moved only from 16/32 to 17/32 (6 fixed, 5 broken). The broken episodes reach the third cube on time and then fail to place it, as the unedited policy also does. On the held-out slots the edit scored 39/64 against 35/64 (17 fixed, 13 broken; exact $p=0.585$). Records are in \texttt{study/results/ctrl/stack3/} (\texttt{FINAL2.md}).

\FloatBarrier
\section{Where to edit: construction, offline checks, and stages}
\label{app:where}
\paragraph{Construction.} Every placement in Section~\ref{sec:where} is built from the same 6 recorded premature-release decisions, with the unedited policy under the explicit wording as the teacher. The two VLM edits are byte copies of the paper's edits, and the gated head edit is a byte copy of the blind editor's submission. Each was checked by its fingerprint. \emph{Action expert.} Keys are the inputs of each layer's MLP down-projection (\texttt{action.blocks.00--17.mlp\_1.linear}, 4,096 features) under the original wording, averaged over the 6 decisions. There is one key per action token (15) and denoising step (10), so 150 keys per layer. Values are the explicit-minus-original residual increment of each layer, with both wordings replayed on the same images, state, and noise. In the action expert, the MLP output is multiplied by an adaptive-norm gate before it is added to the residual stream. Depending on the layer, the gate is below 0.05 for 27--99\% of the dimensions. The update therefore solves the same ridge objective through the gate, $\min_{\Delta W}\lVert g\odot(K\Delta W)-V\rVert^2+\lambda\lVert\Delta W\rVert^2$, which reduces to Eq.~\eqref{eq:edit} when every gate is 1. \emph{Action head, least norm.} Keys are the head inputs after the final norm (150 keys, as above). Values are the explicit wording's velocity minus the original's in the 8 output dimensions the robot uses, and the update has rank at most 8. \emph{VLM + action expert.} The action-expert update is computed on the VLM-edited model and added to it. The VLM edit changes 450,468,837 stored values (13.4\%), the 0.16\% variant 248,057,499 (7.4\%), VLM + action expert 505,694,140 (15.1\%), the action expert 60,378,838 (1.8\%), the least-norm head edit 8,056, and the gated head edit 1,024.

\paragraph{Offline checks.} On the 49 recorded decisions of Appendix~\ref{app:construction} (21 premature releases, 9 correct carries, 19 phase-two decisions), the VLM edit, VLM + action expert, and the gated head edit flip all 21 releases and change no carry or phase-two decision. The action-expert edit leaves 1 release and closes the gripper at 2 of the 7 phase-two approaches to the banana. The least-norm head edit leaves 2 and closes it at 6 of 7. At the first denoising step, the least-norm head edit changes the gripper velocity by $-1.09$ at the premature releases and $-1.10$ at the phase-two approaches, whereas the gated edit changes it by $-1.46$ and $-0.15$. Relative to the weights, the VLM edit's update has a median norm of 0.017 per layer (range 0.005--0.083) and the action-expert update 0.99 (0.04--2.22). At the 6 distilled decisions, the VLM edit reproduces 104--114\% of the explicit wording's change of the action expert's residual stream in every layer. The action-expert edit overshoots (112--126\%), and VLM + action expert comes closest (99--105\%). Applying Eq.~\eqref{eq:edit} to the action expert while ignoring the gate left all 21 releases.

\paragraph{Stages.} Table~\ref{tab:where-stages} breaks the held-out episodes into stages from RoboLab's score and events. The VLM edit places the cube in 60 of 64 episodes and completes the banana step in 92\% of those. The action-expert edit closes the gripper in its first action chunk in 52\% of episodes, and the least-norm head edit in 94\%.

\begin{table}[htbp]
\centering\small
\caption{Where to edit, held-out episodes by stage (64 shared seeds). Banana step: share of the episodes that place the cube and then succeed. First chunk closes: share of episodes whose first action chunk closes the gripper. Wrong grabs: wrong-object grab events per episode.}
\label{tab:where-stages}
\setlength{\tabcolsep}{4pt}
\begin{tabular}{@{}lrrrrr@{}}
\toprule
Edit location & Cube placed & Success & Banana step & First chunk closes & Wrong grabs\\
\midrule
\input{generated/where_stages_table.tex} 
\bottomrule
\end{tabular}
\end{table}

\paragraph{Lanes and records.} An evaluation chunk that ran on a single GPU lane starts its second 16 slots from different initial scenes, and 12 of the 14 held-out chunks did. We therefore rechecked every comparison on the 32 slots never affected. The ordering and every significant difference were unchanged. The transfer and retention runs used fresh slots 94016--94047, with the unedited policy under each task's own instruction as the shared reference. The least-norm head edit, which fixed nothing, and the two remaining VLM placements were not run there. The study used 5.9 GPU-hours. Its pre-registration, scripts, and results are in \texttt{study/results/opus/where/}.

\FloatBarrier
\section{Reproducibility}
\label{app:repro}
\paragraph{Code and data.} The paired outcomes behind every comparison in this paper are in the repository \url{https://github.com/adeeb10abbas/weightsmith}. The paired comparisons are in \texttt{study/results/opus/comparisons.json}, and the four held-out runs, with per-slot outcomes, Wilson intervals, and exact McNemar tests, are in \texttt{study/results/opus/heldout/}. The evidence record \texttt{study/results/opus/EVIDENCE.md} lists every proposal, budget, and cost. The edit scripts are in \texttt{study/results/opus/ctb/}, and the brief and final report of the same-protocol run are in \texttt{study/results/opus/}. The blind editors' briefs, notebooks, scripts, write-ups, and held-out results are in \texttt{study/results/ctrl/}, and the where-to-edit study's pre-registration, scripts, and results in \texttt{study/results/opus/where/}. The command \texttt{python3 paper/scripts/make\_figures.py} regenerates every figure and generated table of this paper from these files.

\paragraph{The released tool.} \texttt{weightsmith distill} (\texttt{weightsmith/distill.py}; usage in \texttt{docs/DISTILL.md}) runs selection, capture, build, and offline check through the editing command. Rebuilt from the stored capture in fresh candidates, both published edits come out identical, value by value. Their fingerprints match the originals, \texttt{63db0c52a51306c9} for the full edit and \texttt{5bf8c4329c1e416a} for the 0.16\% variant (\texttt{docs/DISTILL.md}). The test \texttt{tests/test\_distill\_reproduce.py} repeats the check on the GPU host. The end-to-end run of Section~\ref{sec:tool} is written up step by step in \texttt{docs/TUTORIAL.md}. Its configuration is \texttt{examples/cube\_then\_banana/distill-first-carry.json}, and the operator's held-out score of its edit (candidate \texttt{ctb-first}, fingerprint \texttt{3047901c3502ffea}, slots 95008--95071) is in \texttt{study/results/tutorial/}.

\paragraph{Held-out protocol.} The operator scored frozen edits with \texttt{weightsmith/heldout.py} on slots 90000--90127 (cube then banana) and 90000--90063 (transfer tasks). The first half of the slots ran the unedited policy first and the second half the edited policy first. Editors never saw held-out slots. On cube then banana, the Wilson 95\% intervals are 75.3--88.4\% for the edited and 14.9--29.0\% for the unedited policy. The random-direction control and the held-out explicit-wording arm ran later with \texttt{study/results/opus/controls/ctl\_run.py} on slots 90000--90063 and are paired with the unedited runs recorded there. Table~\ref{tab:halves} splits each held-out run by run order, and every gain appears in both halves. The RoboLab event logs of the cube-then-banana run are summarized per slot in \texttt{paper/figures/src/heldout\_v2mif\_events.json}, which gives the cube-drop counts in Section~\ref{sec:results}. Fig.~\ref{fig:teaser} shows the lowest-numbered fixed start whose unedited run dropped the cube before touching the banana, start 90002. Its frames, gripper commands, and events come from the recorded episodes (\texttt{paper/scripts/build\_teaser\_data.py}).

\begin{table}[htbp]
\centering\small
\caption{Held-out runs split by run order, from the per-slot outcomes. Counts are successes out of the slots in each half. Every gain appears in both halves.}
\label{tab:halves}
\begin{tabular}{@{}llllll@{}}
\toprule
Run & Slots & Order & Unedited & Edited & Fixed / broke\\
\midrule
\input{generated/heldout_halves_table.tex} 
\bottomrule
\end{tabular}
\end{table}

\paragraph{Runtime and editors.} Evaluations used RoboLab on Isaac Sim with the performance preset and 16 environments per evaluation. The original cohort ran on one RTX 3090. The runtime code is identical to commit \texttt{f503fb3} of the repository, and the checkpoint is pinned by \texttt{third\_party/pi05\_droid\_jointpos.sha256}. Claude Opus editors ran headless through Claude Code, with tool access limited to the editing command and file operations. All 730 weight tensors were editable, and the budgets were 1,200 (stacking) or 800 (flat box) practice episodes and 6,000 activation items. Practice slots were 20000--20063 for stacking and 60000--60063 for the flat box and mustard. The cube-then-banana practice runs used slots 20000--20031. Raw episodes, activation captures, and candidate weights remain on the GPU host and are not in the repository.

\paragraph{Cost.} Recorded editor API spend for the Claude Opus cohort was about \$799: \$632 for the stacking editor, \$94 for the first project, and \$73 for the flat-box editors. The same-protocol run cost about \$20. Operator sessions and GPU time were not metered.

\end{document}

%% file: generated/survey_table.tex
Cube into bowl & 16/16 & ---\\
Banana onto plate & 16/16 & ---\\
Butter box onto raisin box & 14/16 & dropped $\times$1, missed $\times$1\\
Stack three cubes & 11/16 & wrong object first $\times$5\\
Block tower, any order & 11/16 & wrong object first $\times$5\\
Stack left bowl on right bowl & 10/16 & wrong object first $\times$5\\
Stack yellow on red & 6/16 & wrong object first $\times$9\\
Pick up eyeglasses & 5/16 & wrong object first $\times$6, grabbed but not lifted $\times$5\\
Mustard into left bin & 3/16 & grabbed then dropped $\times$13\\
Cube then banana into bowl & 2/16 & wrong object first $\times$8, placed but not success $\times$4\\
Cube left of bowl & 1/16 & wrong object first $\times$14\\
Marker in mug & 1/16 & closed but missed $\times$6, mixed\\
Yogurt into bowl & 1/16 & wrong object first $\times$7, mixed\\
Cube right of bowl & 0/16 & wrong object first $\times$16\\
Grab a bagel & 0/16 & wrong object first $\times$7, mixed\\
Pick up flat green box & 0/16 & grabbed but not lifted $\times$10\\

%% file: generated/comparisons_table.tex
\multicolumn{6}{@{}l}{\textit{Language binding (cube then banana)}}\\
Explicit wording (no edit) & --- & 22/32 vs 3/32 & 20 / 1 & +59.4 [+40.6, +78.1] & $2.1\times10^{-5}$\\
Distilled edit, practice & 450,468,837 & 25/32 vs 3/32 & 22 / 0 & +68.8 [+53.1, +84.4] & $4.8\times10^{-7}$\\
Distilled edit, held-out & 450,468,837 & 106/128 vs 27/128 & 81 / 2 & +61.7 [+52.3, +70.3] & $7.2\times10^{-22}$\\
With preservation keys (v2p-f) & --- & 24/32 vs 3/32 & 21 / 0 & +65.6 [+50.0, +81.2] & $9.5\times10^{-7}$\\
0.16\% variant, held-out & 248,057,499 & 102/128 vs 19/128 & 86 / 3 & +64.8 [+55.5, +73.4] & $3.8\times10^{-22}$\\
Random directions, same size, held-out & 502,291,925 & 9/64 vs 13/64 & 6 / 10 & $-$6.2 [$-$18.8, +6.2] & 0.45\\
Explicit wording, held-out (no edit) & --- & 57/64 vs 13/64 & 45 / 1 & +68.8 [+56.2, +79.7] & $1.3\times10^{-12}$\\
\addlinespace
\multicolumn{6}{@{}l}{\textit{Same distilled edit, other tasks}}\\
Cube left of bowl: held-out & 450,468,837 & 36/64 vs 8/64 & 33 / 5 & +43.8 [+28.1, +59.4] & $4.3\times10^{-6}$\\
Cube left of bowl: practice & 450,468,837 & 22/32 vs 3/32 & 20 / 1 & +59.4 [+40.6, +78.1] & $2.1\times10^{-5}$\\
Cube right of bowl: practice & 450,468,837 & 0/32 vs 0/32 & 0 / 0 & 0.0 [0.0, 0.0] & 1\\
Stack three cubes: practice & 450,468,837 & 19/32 vs 14/32 & 9 / 4 & +15.6 [$-$6.2, +37.5] & 0.27\\
Block tower: practice & 450,468,837 & 14/32 vs 15/32 & 7 / 8 & $-$3.1 [$-$28.1, +21.9] & 1\\
Yogurt into bowl: practice & 450,468,837 & 2/32 vs 3/32 & 2 / 3 & $-$3.1 [$-$15.6, +9.4] & 1\\
Grab a bagel: practice & 450,468,837 & 1/32 vs 0/32 & 1 / 0 & +3.1 [0.0, +9.4] & 1\\
Marker in mug: practice & 450,468,837 & 1/32 vs 0/32 & 1 / 0 & +3.1 [0.0, +9.4] & 1\\
Pick up eyeglasses: held-out & 450,468,837 & 44/64 vs 22/64 & 31 / 9 & +34.4 [+17.2, +51.6] & $6.8\times10^{-4}$\\
Pick up eyeglasses: practice, two runs & 450,468,837 & 46/64 vs 26/64 & 28 / 8 & +31.2 [+14.1, +46.9] & 0.0012\\
Mustard into bin: practice & 450,468,837 & 12/32 vs 7/32 & 11 / 6 & +15.6 [$-$9.4, +40.6] & 0.33\\
Stacking: practice & 450,468,837 & 17/32 vs 14/32 & 9 / 6 & +9.4 [$-$15.6, +34.4] & 0.61\\
Stack bowls: practice & 450,468,837 & 21/32 vs 19/32 & 8 / 6 & +6.2 [$-$15.6, +28.1] & 0.79\\
Butter: edit vs first unedited run & 450,468,837 & 10/16 vs 14/16 & 0 / 4 & $-$25.0 [$-$50.0, $-$6.2] & 0.12\\
Butter: preservation keys vs first run & 450,468,837 & 10/16 vs 14/16 & 0 / 4 & $-$25.0 [$-$50.0, $-$6.2] & 0.12\\
Cube into bowl: practice & 450,468,837 & 16/16 vs 16/16 & 0 / 0 & 0.0 [0.0, 0.0] & 1\\
Banana onto plate: practice & 450,468,837 & 16/16 vs 16/16 & 0 / 0 & 0.0 [0.0, 0.0] & 1\\
\addlinespace
\multicolumn{6}{@{}l}{\textit{Perception: lookalike target (stack yellow on red)}}\\
Attention gain on identity axis (v1a1) & 30,400,000 & 21/64 vs 28/64 & 9 / 16 & $-$10.9 [$-$26.6, +4.7] & 0.23\\
VLM key/value gain (dkv2) & 12,000,000 & 21/64 vs 28/64 & 11 / 18 & $-$10.9 [$-$26.6, +6.2] & 0.26\\
Key/value gain + head rewire (combo1) & 12,200,000 & 21/64 vs 28/64 & 9 / 16 & $-$10.9 [$-$26.6, +4.7] & 0.23\\
Same weights as combo1, second run & 12,200,000 & 21/64 vs 28/64 & 8 / 15 & $-$10.9 [$-$25.0, +3.1] & 0.21\\
Astra A1 p1 (4,412 values) & 4,412 & 12/32 vs 11/32 & 7 / 6 & +3.1 [$-$18.8, +25.0] & 1\\
Astra A2 p1 (1,543 values) & 1,543 & 13/32 vs 11/32 & 9 / 7 & +6.2 [$-$18.8, +31.2] & 0.80\\
Astra A2 p3 (791 values) & 791 & 13/32 vs 11/32 & 7 / 5 & +6.2 [$-$15.6, +28.1] & 0.77\\
Astra fixed16 p1 (6,225 values) & 6,225 & 11/32 vs 17/32 & 5 / 11 & $-$18.8 [$-$40.6, +6.2] & 0.21\\
Claude, Astra protocol p1 (64 rows) & --- & 16/32 vs 12/32 & 8 / 4 & +12.5 [$-$9.4, +34.4] & 0.39\\
Claude, Astra protocol p2 (35,210 values) & 35,210 & 11/32 vs 12/32 & 6 / 7 & $-$3.1 [$-$25.0, +18.8] & 1\\
\addlinespace
\multicolumn{6}{@{}l}{\textit{Missing competence (flat box)}}\\
10 edits (readout, rows, flips, biases) & --- & 0/16 each vs 0/64 & 0 / 0 & 0.0 & ---\\
\addlinespace
\multicolumn{6}{@{}l}{\textit{Contact physics (mustard)}}\\
Gripper bias (1 value) & 1 & 13/64 vs 13/64 & 10 / 10 & 0.0 [$-$14.1, +14.1] & 1\\
\addlinespace
\multicolumn{6}{@{}l}{\textit{Gripper commitment}}\\
Stacking: gripper bias (1 value) & 1 & 27/64 vs 27/64 & 12 / 12 & 0.0 [$-$15.6, +15.6] & 1\\
Pick up eyeglasses: gripper bias (1 value) & 1 & 13/32 vs 12/32 & 7 / 6 & +3.1 [$-$18.8, +25.0] & 1\\
Stack bowls: gripper bias (1 value) & 1 & 18/32 vs 19/32 & 6 / 7 & $-$3.1 [$-$25.0, +18.8] & 1\\
Butter: gripper bias (1 value) & 1 & 14/16 vs 14/16 & 2 / 2 & 0.0 [$-$25.0, +25.0] & 1\\
Cube into bowl: gripper bias (1 value) & 1 & 16/16 vs 16/16 & 0 / 0 & 0.0 [0.0, 0.0] & 1\\
Banana onto plate: gripper bias (1 value) & 1 & 16/16 vs 16/16 & 0 / 0 & 0.0 [0.0, 0.0] & 1\\
Stacking: runtime threshold 0.4 (Astra study) & --- & 36/64 vs 23/64 & 21 / 8 & +20.3 [+4.7, +35.9] & 0.024\\
Seven tasks pooled: gripper bias (1 value) & 1 & 117/240 vs 117/240 & 37 / 37 & 0.0 [$-$7.1, +7.1] & 1\\
\addlinespace
\multicolumn{6}{@{}l}{\textit{Unchanged weights (noise controls)}}\\
Stacking: unchanged rerun (Astra study) & --- & 23/64 vs 26/64 & 17 / 20 & $-$4.7 [$-$23.4, +14.1] & 0.74\\
Stacking: extra unchanged rerun (Astra study) & --- & 35/64 vs 26/64 & 21 / 12 & +14.1 [$-$3.1, +31.2] & 0.16\\
Stacking: unchanged rerun, fixed16 (Astra study) & --- & 27/64 vs 24/64 & 16 / 13 & +4.7 [$-$12.5, +21.9] & 0.71\\
Butter: unedited rerun, same seeds & --- & 8/16 vs 14/16 & 0 / 6 & $-$37.5 [$-$62.5, $-$12.5] & 0.031\\

%% file: generated/gripper_table.tex
Mustard into bin & 64 & 13 & 13 & 10 / 10\\
Stack yellow on red & 64 & 27 & 27 & 12 / 12\\
Pick up eyeglasses & 32 & 12 & 13 & 7 / 6\\
Stack bowls & 32 & 19 & 18 & 6 / 7\\
Butter on raisin box & 16 & 14 & 14 & 2 / 2\\
Cube into bowl & 16 & 16 & 16 & 0 / 0\\
Banana onto plate & 16 & 16 & 16 & 0 / 0\\
\midrule
Total & 240 & 117 & 117 & 37 / 37\\

%% file: generated/transfer_table.tex
Pick up eyeglasses, held-out & 64 & 22 & 44 & 31 / 9 & +34.4 [+17.2, +51.6] & $6.8\times10^{-4}$\\
Pick up eyeglasses, two practice runs & 64 & 26 & 46 & 28 / 8 & +31.2 [+14.1, +46.9] & 0.0012\\
Cube left of bowl, held-out & 64 & 8 & 36 & 33 / 5 & +43.8 [+28.1, +59.4] & $4.3\times10^{-6}$\\
Cube left of bowl, practice & 32 & 3 & 22 & 20 / 1 & +59.4 [+40.6, +78.1] & $2.1\times10^{-5}$\\
\addlinespace
Cube right of bowl & 32 & 0 & 0 & 0 / 0 & 0.0 [0.0, 0.0] & 1\\
Stack three cubes & 32 & 14 & 19 & 9 / 4 & +15.6 [$-$6.2, +37.5] & 0.27\\
Block tower & 32 & 15 & 14 & 7 / 8 & $-$3.1 [$-$28.1, +21.9] & 1\\
Yogurt into bowl & 32 & 3 & 2 & 2 / 3 & $-$3.1 [$-$15.6, +9.4] & 1\\
Grab a bagel & 32 & 0 & 1 & 1 / 0 & +3.1 [0.0, +9.4] & 1\\
Marker in mug & 32 & 0 & 1 & 1 / 0 & +3.1 [0.0, +9.4] & 1\\
Pick up flat box & 32 & 0 & 0 & 0 / 0 & 0.0 [0.0, 0.0] & 1\\
Mustard into bin & 32 & 7 & 12 & 11 / 6 & +15.6 [$-$9.4, +40.6] & 0.33\\
Stack yellow on red & 32 & 14 & 17 & 9 / 6 & +9.4 [$-$15.6, +34.4] & 0.61\\
Stack bowls & 32 & 19 & 21 & 8 / 6 & +6.2 [$-$15.6, +28.1] & 0.79\\
Butter on raisin box, vs first unedited run & 16 & 14 & 10 & 0 / 4 & $-$25.0 [$-$50.0, $-$6.2] & 0.12\\
Cube into bowl & 16 & 16 & 16 & 0 / 0 & 0.0 [0.0, 0.0] & 1\\
Banana onto plate & 16 & 16 & 16 & 0 / 0 & 0.0 [0.0, 0.0] & 1\\
\addlinespace
Butter, unedited rerun vs first unedited run & 16 & 14 & 8 & 0 / 6 & $-$37.5 [$-$62.5, $-$12.5] & 0.031\\

%% file: generated/where_stages_table.tex
Unedited & 17/64 & 5/64 & 29\% & 2\% & 2.89\\
VLM, full edit & 60/64 & 55/64 & 92\% & 2\% & 0.19\\
VLM, 0.16\% variant & 50/64 & 41/64 & 82\% & 2\% & 0.80\\
VLM + action expert & 56/64 & 43/64 & 77\% & 2\% & 0.86\\
Action expert & 27/64 & 20/64 & 74\% & 52\% & 4.52\\
Action head, least norm & 0/64 & 0/64 & --- & 94\% & 0.00\\
Action head, gated (blind editor) & 40/64 & 30/64 & 75\% & 9\% & 1.55\\

%% file: generated/heldout_halves_table.tex
Cube then banana, distilled edit & 90000--90063 & unedited first & 13/64 & 50/64 & 38 / 1\\
 & 90064--90127 & edited first & 14/64 & 56/64 & 43 / 1\\
Cube then banana, 0.16\% variant & 90000--90063 & unedited first & 9/64 & 54/64 & 46 / 1\\
 & 90064--90127 & edited first & 10/64 & 48/64 & 40 / 2\\
Pick up eyeglasses, distilled edit & 90000--90031 & unedited first & 8/32 & 19/32 & 17 / 6\\
 & 90032--90063 & edited first & 14/32 & 25/32 & 14 / 3\\
Cube left of bowl, distilled edit & 90000--90031 & unedited first & 3/32 & 15/32 & 15 / 3\\
 & 90032--90063 & edited first & 5/32 & 21/32 & 18 / 2\\